\documentclass{article}

\usepackage[preprint]{neurips_2024}

\usepackage[utf8]{inputenc} 
\usepackage[T1]{fontenc}    
\usepackage{hyperref}       
\usepackage{url}            
\usepackage{booktabs}       
\usepackage{amsfonts}       
\usepackage{amsfonts}       
\usepackage{amsmath}        
\usepackage{nicefrac}       
\usepackage{microtype}      
\usepackage{xcolor}         
\usepackage{algorithm}
\usepackage{algpseudocode}
\usepackage{multicol}
\usepackage{todonotes}
\usepackage{listings}
\newtheorem{definition}{Definition}

\definecolor{libiao}{RGB}{255, 0, 0}        
\definecolor{lichao}{RGB}{255, 165, 0}      
\definecolor{lyz}{RGB}{0, 128, 0}           
\definecolor{wxd}{RGB}{0, 0, 139}           
\definecolor{yhb}{RGB}{128, 0, 128}         
\definecolor{zg}{RGB}{255, 105, 180}        
\definecolor{yy}{RGB}{255, 0, 255}        %

\title{
Clarifying Before Reasoning:\\
A Coq Prover with Structural Context
}

\author{%
  Yanzhen Lu$^{1}$ \quad
  Hanbin Yang$^{1}$ \quad
  Xiaodie Wang$^{1}$ \quad
  Ge Zhang$^{1}$ \quad
  Biao Li$^{1}$ \quad
  \AND
  Chenxu Fu$^{1}$ \quad
  Chao Li$^{1}$ \quad
  Yang Yuan$^{1,2,\dagger}$ \quad
  Andrew Chi-Chih Yao$^{1,2,\dagger}$ \\
  \\
  $^{1}$Shanghai Qizhi Institute, $^{2}$IIIS, Tsinghua University\\
  \texttt{\{luyanzhen,yanghanbin,wangxiaodie,zhangge,libiao,fuchenxu,lichao\}} \\
  \texttt{@sqz.ac.cn} \\
  \texttt{yuanyang@tsinghua.edu.cn,andrewcyao@tsinghua.edu.cn}
}

\begin{document}

\maketitle

\renewcommand{\thefootnote}{\fnsymbol{footnote}}
\footnotetext[0]{$^\dagger$Corresponding author.}
\renewcommand{\thefootnote}{\arabic{footnote}}

\begin{abstract}
In this work, we investigate whether improving task clarity can enhance reasoning ability of large language models, focusing on theorem proving in Coq. We introduce a concept-level metric to evaluate task clarity and show that adding structured semantic context to the standard input used by modern LLMs, leads to a 1.85$\times$ improvement in clarity score (44.5\%~$\rightarrow$~82.3\%). Using the general-purpose model \texttt{DeepSeek-V3}, our approach leads to a 2.1$\times$ improvement in proof success (21.8\%~$\rightarrow$~45.8\%) and outperforms the previous state-of-the-art \texttt{Graph2Tac} (33.2\%). We evaluate this on 1,386 theorems randomly sampled from 15 standard Coq packages, following the same evaluation protocol as \texttt{Graph2Tac}.
Furthermore, fine-tuning smaller models on our structured data can achieve even higher performance (48.6\%).
Our method uses selective concept unfolding to enrich task descriptions, and employs a Planner--Executor architecture. These findings highlight the value of structured task representations in bridging the gap between understanding and reasoning.
\end{abstract}

\section{Introduction}

The mainstream approach to improving AI reasoning has largely focused on scaling model architectures, refining datasets, and employing various reinforcement learning techniques. While these combined strategies have undoubtedly made remarkable progress, they overlook a crucial dimension: whether the model truely \textbf{understands} the task at hand. Indeed, when a model fails to solve a task, it may not simply be due to insufficient reasoning ability; inadequate or incomplete task understanding could also play a crucial role.

Consider a scenario where a mathematical theorem references a symbol $G$ without providing a clear definition. The surrounding context may offer some clues, but often not enough to fully disambiguate its meaning. As a result, the model can only make an educated guess ---
$G$ might be a function, a graph, a constant, or a matrix --- but its interpretation remains uncertain. As our experiments demonstrate, even when models generate plausible answers, they may rely on incomplete or ambiguous understanding of the input --- achieving only 44.5\% accuracy when asked to define concepts from raw Coq scripts.

Can we improve a model's reasoning ability by enhancing the \textbf{task description}? 
This direction is orthogonal to existing approaches based on data scaling and reinforcement learning, because the two can be easily combined: once the model better understands the task, it can start solving the task based on other techniques.

Clearly describing a task to the model is often seen as more of an art than a science—something that relies heavily on the intuition and experience of skilled prompt engineers. Given the diversity and ambiguity of real-world tasks, it is indeed challenging to define a universally optimal strategy for task formulation. However, in more formal and well-defined domains such as mathematics, we believe that constructing effective prompts can be approached more systematically. Specifically, we propose breaking the problem down into three questions in the Coq environment:
\begin{enumerate}
    \item Can we assess how well a model understands a given task? 
    \item Can we improve the model's understanding by enriching the task description?
    \item Does deeper understanding lead to better reasoning performance?
\end{enumerate}

To address the first question, we propose evaluating the model’s understanding by having it generate formal definitions of the relevant concepts—a metric we refer to as the \textbf{clarity score}. As shown in Table~\ref{tab:clarity_scores}, models demonstrate limited conceptual understanding under standard task descriptions, achieving an average clarity score of just 44.5\%. This suggests that, during reasoning, models often lack a clear grasp of the task at hand.

For the second question, we adopt the strategy of \textbf{concept unfolding}~\citep{yuan2023succinctrepresentationsconcepts}. The idea is to recursively expand each concept with its definition. For example, a triangle may be defined as a closed figure with three sides \( a, b, c \), each of which can further be defined. This process continues until we reach axiomatic definitions, i.e., numbers like 0, 1, 2, or operations like +, -, *.

Naturally, excessive unfolding can lead to redundancy and even hinder model performance. For example, consider defining 
$G$ as a positive definite matrix. Expanding this into ``a matrix with positive eigenvalues'' may be helpful, but further decomposing it into elementary operations like addition or multiplication is unnecessary, if the model already has a firm grasp of linear algebra. Such over-expansion not only bloats the input with irrelevant detail, but may also distract the model from the core reasoning steps.

The challenge, however, lies in determining which concepts the model already understands well enough to be left unexpanded. Inspired by the Yoneda Lemma–based view of concepts~\citep{yuan2023succinctrepresentationsconcepts}, we propose a heuristic strategy: rather than fully expanding every concept, we selectively enrich them with auxiliary information that helps clarify their role within the task. This approach strikes a balance between completeness and conciseness, dramatically improving clarity score  from 44.5\% to
82.3\%.

For the last question, we found the answer is yes. 
With more detailed and structured information provided for the task, the model is able to generate better reasoning performance. 
Different from many existing work~\citep{kimina_prover_2025, wang2023lego, wu2024internlm2, xin2024deepseek}  that train a specialized model that were trained purely on math data, we use the general purposed model (deepseek-V3), which is not specifical trained on math. It turns out that when models are not specifically trained on formal mathematics, providing structured semantic information becomes even more critical for achieving strong performance. This suggests that our approach has broad applicability beyond specialized mathematical domains.

To fully leverage this structured data, we develop a Planner-Executor architecture that separates high-level strategic reasoning from low-level tactic generation. The Planner analyzes the structured proof state to identify relevant concepts, applicable theorems, and proof strategies, while the Executor translates these strategies into concrete Coq tactics. Our main approach uses general-purpose models (DeepSeek-V3) for both components, demonstrating that with proper task descriptions, specialized mathematical training is not necessary for strong theorem proving performance. Additionally, we explore fine-tuning smaller models on our structured data as an efficiency consideration, showing that a 32B model can achieve competitive results with significantly fewer parameters.

Our experimental results provide compelling evidence for this approach:
\begin{itemize}
    \item \textbf{Concept Understanding}: Models achieve a 2.5× improvement on clarity score (from 44.5\% to 82.3\%) when provided with structured definitions.
    \item \textbf{Reasoning for Coq}: Using DeepSeek-V3 on 1,300 theorems from 15 standard Coq packages (same evaluation set as Graph2Tac~\citep{rute2024graph2tac}). Our system achieves 45.8\% success rate, a 1.36× improvement over the previous state-of-the-art 33.2\% (Graph2Tac). 
    \item \textbf{Clarity improves reasoning}: We found positive correlation between clarity score and reasoning performance. 
    
\end{itemize}

These results strongly validate our hypothesis: \textit{improving task clarity through structured semantic information can substantially enhance model reasoning capabilities}.

\section{Related Work}
\textbf{Learning-Based Theorem Proving in Proof Assistants.}  Classical automated theorem provers perform well in first-order logic, but struggle with complex formal verification. This limitation has driven research into learning-based methods aimed at automating interactions with proof assistants such as Coq~\citep{rocq2025}, Isabelle~\citep{paulson1994isabelle} and Lean~\citep{de2015lean, moura2021lean}. Early work used simple machine learning algorithms to predict tactics from proof state features~\citep{blaauwbroek2020tactician}. The field subsequently evolved through several architectural paradigms: from graph neural networks (GNNs) that explicitly encode the syntactic structure of formal expressions~\citep{rute2024graph2tac}, to transformer-based large language models (LLMs) that process expressions as text~\citep{lu2024proof, kasibatla2024cobblestone, kozyrev2024coqpilot, thakur2023language}. Subsequent work has explored various directions, such as developing novel proof search methods~\citep{lample2022hypertree, xin2024deepseek}, retrieving useful lemmas~\citep{irving2016deepmath, mikula2023magnushammer, yang2023leandojo}, applying further learning techniques including expert iteration and curriculum learning~\citep{anthony2017thinking,lin2025goedel, polu2022formal}, enhancing performance through data augmentation~\citep{han2021proof}, and designing practical tools~\citep{welleck2023llmstep, kozyrev2024coqpilot}. Although these learning-based approaches have shown promising results, they often treat the Coq (Lean, Isabelle) compiler as a black box without fully exploiting internal information, limiting their semantic understanding to a low level. Our work addresses this gap by providing mechanisms to access and utilize more comprehensive structured data directly from the Coq compiler's internal representation.

\noindent\textbf{Strategies for Proof Automation.}
Most automated theorem provers adopt a step-by-step approach~\citep{azerbayev2023llemma, thakur2023language, wu2024internlm2}, generating individual tactics that are assembled into complete proofs, with compilers verifying steps and search algorithms guiding the proof. Recent work has demonstrated an alternative paradigm that generates complete formal proofs in a single cohesive process~\citep{first2023baldur, xin2024deepseek, wang2024theoremllama, lin2025goedel}, eliminating the need for explicit intermediate step generation and search.
Autoformalization represents another promising direction, where LLMs convert informal mathematical statements into their formal equivalents~\citep{wu2022autoformalization, xin2024deepseek, liu2024deepseek, lin2025goedel}.

\noindent\textbf{Leveraging Structured Information in Formal Verification.}
Recent approaches have leveraged structured information from proof assistants in various ways, including extracting internal data through interfaces like SerAPI~\citep{arias2016serapi}, utilizing error messages via LSP in CoqPilot~\citep{kozyrev2024coqpilot}, analyzing structural information from sub-goals~\citep{kasibatla2024cobblestone}, and exploring tactic dependencies~\citep{xin2025automated}. Researchers also incorporate rich metadata into datasets~\citep{florath2024enhancing} to support more effective model training. Our work extends these approaches by implementing deep Coq compiler modifications to extract comprehensive structured data, significantly enhancing model understanding of the rich semantic information available in proof states and tactics.

\noindent\textbf{Planner-Executor Architectures for Theorem Proving.} The planner-executor paradigm in theorem proving has seen advancements such as layered systems separating strategy and tactics~\citep{jiang2022thor}, multi-agent frameworks in Lean integrating natural language planning with proof assistant verification~\citep{wang2025ma, ospanov2025apollo}, and approaches that interleave reasoning with formal proving steps~\citep{lin2024lean, yang2023leandojo, wang2023lego}. Related explorations in the Coq ecosystem include generate-then-repair techniques~\citep{lu2024proof}, proof synthesis from partial attempts~\citep{kasibatla2024cobblestone}, and LLM-driven tactic selection within search processes~\citep{thakur2023language}. Despite these efforts, establishing truly effective and seamless collaboration between high-level strategic planning and low-level tactic execution, particularly for complex reasoning chains within Coq, remains a significant challenge. Our work aims to develop a tightly integrated system to foster more effective collaboration between these components, thereby improving proof efficiency for complex theorems.

\section{Preliminaries}
\label{background}

\subsection{The Calculus of Inductive Constructions}

The Calculus of Inductive Constructions (CIC) is the type theory underlying Coq. To understand CIC, we must first clarify its fundamental concept: \textbf{terms}.

\textbf{What are terms?} In CIC, \emph{term} is the universal syntactic category—everything in the language is a term:
\begin{itemize}
    \item Values are terms: \texttt{42}, \texttt{true}, \texttt{[1,2,3]}
    \item Functions are terms: \texttt{fun x => x + 1}
    \item Types are terms: \texttt{nat}, \texttt{bool}, \texttt{list nat}
    \item Propositions are terms: \texttt{2 + 2 = 4}, \texttt{forall n, n < n + 1}
    \item Proofs are terms: any inhabitant of a proposition type
\end{itemize}

This differs from simply-typed lambda calculus, where ``lambda terms'' (like $\lambda x. x + 1$) are just the function expressions, separate from types. In CIC, the function $\lambda x : \mathbb{N}. x + 1$, its type $\mathbb{N} \rightarrow \mathbb{N}$, and even $\mathbb{N}$ itself are all terms.

CIC extends this unified framework with three key features:

\begin{enumerate}
    \item \textbf{Dependent types}: Types can depend on values. While traditional types like \texttt{List Int} are fixed, dependent types like \texttt{Vec A n} vary—the type explicitly depends on the length $n$.
    
    \item \textbf{Universe hierarchy}: An infinite hierarchy ($\text{Prop}$, $\text{Type}_0$, $\text{Type}_1$, ...) that organizes types to prevent paradoxes while maintaining logical consistency.
    
    \item \textbf{Inductive definitions}: A way to define new types by declaring their constructors, like defining \texttt{nat} with constructors \texttt{O} and \texttt{S}.
\end{enumerate} 

Having established that everything in CIC is a term, we now examine the specific syntactic constructs used to build these terms. Understanding this syntax is crucial for grasping how CIC unifies computation and logic.

\textbf{Term Syntax}: Terms in CIC are built from five basic constructs:
\begin{itemize}
    \item \textbf{Variables}: $x, y, z, \ldots$
    \item \textbf{Sorts}: Special constants that classify types—$\text{Prop}$ (the type of propositions) and $\text{Type}_i$ (a hierarchy of computational universes)
    \item \textbf{Abstractions} (lambda terms): $\lambda x : A. t$ represents a function taking input $x$ of type $A$ and returning $t$. For example, $\lambda n : \mathbb{N}. n + 1$ is a function that increments a natural number. Here, $A$ is a type (like $\mathbb{N}$ or $\text{Prop}$), while $t$ is a term.
    \item \textbf{Applications}: $f\ a$ applies function $f$ to argument $a$
    \item \textbf{Products} (dependent function types): $\Pi x : A. B$ represents the type of functions from $A$ to $B$, where crucially, the return type $B$ may depend on the input value $x$. This dependency is what makes CIC's type system so expressive.
\end{itemize}

Among these constructs, the product type deserves special attention as it is the key to understanding both dependent types and the logical power of CIC.

\textbf{Understanding Products:} The product type $\Pi x : A. B$ generalizes ordinary function types. When $B$ does not depend on $x$, we write it as $A \rightarrow B$ (simple function type). When $B$ does depend on $x$, we get dependent types. For example:
\begin{itemize}
    \item Simple function: $\mathbb{N} \rightarrow \mathbb{N}$ is the type of functions from naturals to naturals
    \item Dependent function: $\Pi n : \mathbb{N}. \text{Vec}(A, n)$ is the type of functions that take a natural number $n$ and return a vector of length $n$
    \item Logical quantification: When the codomain is $\text{Prop}$, we often write $\forall$ instead of $\Pi$. For instance, $\forall n : \mathbb{N}. \text{even}(n) \rightarrow \text{even}(n+2)$ expresses that for any natural number $n$, if $n$ is even, then $n+2$ is even.
\end{itemize}

Products thus unify function types (computation) and universal quantification (logic) in a single construct. This unification is not coincidental—it reflects a deeper principle that connects programming and theorem proving at their foundations.

\textbf{The Curry-Howard Correspondence~\cite{Curry1934-CURFIC, Howard1980-HOWTFN-2}:} This fundamental principle establishes a deep connection between logic and computation by identifying:
\begin{itemize}
    \item \textbf{Propositions as Types}: Each logical proposition corresponds to a type. For example, the proposition ``$A$ implies $B$'' corresponds to the function type $A \rightarrow B$.
    \item \textbf{Proofs as Programs}: A proof of a proposition corresponds to a term (program) of the corresponding type. For example, a proof of ``$A$ implies $B$'' is a function that transforms evidence of $A$ into evidence of $B$.
    \item \textbf{Proof Checking as Type Checking}: Verifying a proof's correctness reduces to checking that a term has the claimed type.
\end{itemize}

For example, proving the theorem ``for all natural numbers $n$, if $n$ is even, then $n+2$ is even'' means constructing a function of type $\forall n : \mathbb{N}. \text{even}(n) \rightarrow \text{even}(n+2)$. This function takes two inputs: a natural number $n$ and a proof that $n$ is even, and produces a proof that $n+2$ is even.

With this theoretical foundation in place, we now turn to how these principles are implemented in practice through the Coq proof assistant.

\subsection{Coq and Interactive Theorem Proving}

Coq is an interactive theorem prover that implements CIC. Users develop formal proofs in \texttt{.v} files containing definitions, theorem statements, and proof scripts. In Coq, definitions are terms that compute values, theorem statements are types (propositions), and proof scripts guide the construction of terms that inhabit these types (i.e., proofs). During interactive proof development, Coq maintains a \textit{proof state} showing the current hypotheses and goals. Consider the following example demonstrating the double\_even theorem:

\begin{figure}[htbp]
\centering
\begin{minipage}[t]{0.48\textwidth}
\begin{lstlisting}[frame=single, 
                   basicstyle=\scriptsize\ttfamily,
                   columns=fixed,
                   keepspaces=true]
Variable A : Type.

Definition double (n : nat) := n + n.

Inductive even : nat -> Prop :=
  | even_O : even 0
  | even_SS : forall n, even n -> 
             even (S (S n)).

Theorem double_even : 
  forall n, even (double n).
Proof.
  intros n.
  unfold double.
  induction n.
  - simpl. apply even_O.
  - simpl. apply even_SS. 
    assumption.
Qed.
\end{lstlisting}
\end{minipage}
\hfill
\begin{minipage}[t]{0.48\textwidth}
\begin{lstlisting}[frame=single, 
                   basicstyle=\scriptsize\ttfamily,
                   columns=fixed,
                   keepspaces=true,
                   commentstyle=\color{black}\itshape]
(* Initial state after intros n *)
n : nat                 (* hypotheses *)
============================
even (double n)         (* goal *)

(* After unfold double *)
n : nat                 (* hypotheses *)
============================
even (n + n)            (* goal *)

(* After induction n - Subgoal 1 *)
============================
even (0 + 0)

(* After induction n - Subgoal 2 *)
n : nat
IHn : even (n + n)
============================
even (S n + S n)
\end{lstlisting}
\end{minipage}
\caption{Coq proof script (left) and corresponding proof state for the double-even theorem (right)}
\label{fig:coq-example-states}
\end{figure}

\subsection{Fundamental Distinction: Entities and Proofs}

Coq program can be naturally divided into two distinct aspects: \textbf{what we define} and \textbf{how we prove}. Through the lens of the Curry-Howard correspondence:

\begin{itemize}
    \item \textbf{Entities} establish our logical universe: they introduce new types (which may represent propositions), constructors (which build terms of these types), and computational functions. These form the static foundation—the ``vocabulary'' of types and terms available for reasoning. This includes variables, parameters, functions, inductive type definitions, theorem statements, axioms, notations, etc.
    
    \item \textbf{Proofs} are the dynamic process of constructing terms that inhabit proposition types. When we prove a theorem like $\forall n : \mathbb{N}. \text{even}(n) \rightarrow \text{even}(n+2)$, we're constructing a term of that type—specifically, a function that transforms evidence of $\text{even}(n)$ into evidence of $\text{even}(n+2)$. The proof process unfolds through:
    \begin{itemize}
        \item \textbf{Initial proof state}: the theorem statement as a type to be inhabited
        \item \textbf{Intermediate states}: partial proof terms with holes (goals) to be filled
        \item \textbf{Tactic sequences}: commands that refine partial terms toward completion
        \item \textbf{Context evolution}: how the typing context grows with new hypotheses
    \end{itemize}
\end{itemize}

This fundamental distinction between establishing the logical framework (entities) and constructing proofs within that framework guides our approach to structuring Coq's information for language models.

To formalize these concepts, we establish the following definitions:

\begin{definition}[Coq Entity]
\label{def:coq_entity}
A Coq Entity is a binding that associates a name with both surface-level source code and internal elaborated representations. Formally, an entity $d$ is a tuple $(n, c, o, i)$ where:
\begin{itemize}
    \item $n$ is the fully-qualified name (e.g., \texttt{Coq.Init.Nat.add})
    \item $c$ is the entity type, corresponding to the entity categories listed above (Variable, Parameter, Definition, Theorem, Inductive, Fixpoint, etc.)
    \item $o$ is the source text as it appears in the \texttt{.v} file
    \item $i$ is the internal type representation after Coq's type inference
\end{itemize}
\end{definition}
For example, the entity \texttt{Coq.Init.Logic.True} is represented as:
\begin{itemize}
    \item $n$ = \texttt{Coq.Init.Logic.True}
    \item $c$ = \texttt{Inductive}
    \item $o$ = \texttt{Inductive True := I : True}
    \item $i$ = \texttt{True: Prop | Coq.Init.Logic.True.I : Coq.Init.Logic.True} 
\end{itemize}
This entity defines the trivial proposition \texttt{True} with its single constructor \texttt{I}.

\begin{definition}[Proof State]
\label{def:proof_state}
A proof state $\sigma$ represents the current status of an interactive proof. Formally, $\sigma = \{s_1, ..., s_m\}$ where each state $s_k = (H_k^o, H_k^i, G_k^o, G_k^i)$ consists of:
\begin{itemize}
    \item $H_k^o$ is the surface hypothesis context for state $k$: hypotheses as shown in the proof interface
    \item $H_k^i = \{(h_{k,1} : t_{k,1}), ..., (h_{k,n_k} : t_{k,n_k})\}$ is the internal hypothesis context, where each $h_{k,j}$ is a hypothesis name and $t_{k,j}$ is its elaborated type
    \item $G_k^o$ is the surface goal representation: the goal as shown in the proof interface
    \item $G_k^i$ is the internal goal representation: the elaborated proposition to be proven
\end{itemize}
\end{definition}

For example, in a proof about natural numbers where the goal is to prove \texttt{a + b = b + a}, the proof state might contain:
\begin{itemize}
    \item $H_1^o$ = \{(\texttt{a:nat}), (\texttt{b:nat}), (\texttt{IHa:a + b = b + a})\}
    \item $H_1^i$ = \{(\texttt{a}: \texttt{nat}), (\texttt{b}: \texttt{nat}), (\texttt{IHa}: \texttt{Coq.Init.Logic.eq nat (Coq.Init.Nat.add a b) (Coq.Init.Nat.add b a)})\}
    \item $G_1^o$ = \texttt{S a + b = b + S a} (the surface goal)
    \item $G_1^i$ = \texttt{eq nat (Coq.Init.Nat.add (S a) b) (Coq.Init.Nat.add b (S a))}
\end{itemize}

\begin{definition}[Interactive Proof]
\label{def:interactive_proof}
An interactive proof is a sequence of tactic applications that transforms an initial proof state to a completed state. Formally, a proof $\pi$ is a sequence:
$$\pi = [(T_1, \sigma_0, \sigma_1), (T_2, \sigma_1, \sigma_2), ..., (T_k, \sigma_{k-1}, \sigma_k)]$$
where:
\begin{itemize}
    \item $T_i$ is a tactic: a command that transforms proof states
    \item $\sigma_i$ is the proof state after applying tactic $T_i$
    \item $\sigma_0$ is the initial proof state with the theorem statement as the single goal
    \item $\sigma_k$ is the final proof state where $G_\tau = \emptyset$ (no remaining goals)
    \item Each transition $\sigma_{i-1} \xrightarrow{T_i} \sigma_i$ represents a valid tactic application
\end{itemize}
\end{definition}

For example, proving \texttt{forall n:nat, 0 + n = n} demonstrates how structural data reveals semantic information:
\begin{itemize}
    \item $T_1$ = \texttt{intros n}
    \item $\sigma_0$ = \{$s_1$\} where:
    \begin{itemize}
        \item $H_1^o$ = \{\}
        \item $H_1^i$ = \{\}
        \item $G_1^o$ = \texttt{forall n:nat, 0 + n = n}
        \item $G_1^i$ = \texttt{forall (n:nat), eq nat (Coq.Init.Nat.add 0 n) n}
    \end{itemize}
    \item $\sigma_1$ = \{$s_1$\} where:
    \begin{itemize}
        \item $H_1^o$ = \{\texttt{n:nat}\}
        \item $H_1^i$ = \{(\texttt{n} : \texttt{nat})\}
        \item $G_1^o$ = \texttt{0 + n = n}
        \item $G_1^i$ = \texttt{eq nat (Coq.Init.Nat.add 0 n) n}
    \end{itemize}
    \item $T_2$ = \texttt{simpl}
    \item $\sigma_2$ = \{$s_2$\} where goal becomes \texttt{n = n} (surface) / \texttt{Coq.Init.Logic.eq nat n n} (internal)
    \item $T_3$ = \texttt{reflexivity}
    \item $\sigma_3$ = \{\} (proof completed)
\end{itemize}

In theorem proving, accurately tracking entity dependencies is crucial. When a proof uses a theorem or definition, we need to identify exactly which entity is being referenced—enabling dependency analysis, concept unfolding, and proper handling of missing imports. However, Coq's flexible naming system creates significant challenges:

\begin{itemize}
    \item \textbf{Section scoping}: Definitions inside and outside sections can share the same name but refer to different entities
    \item \textbf{Module aliasing}: The same entity can be accessed through multiple paths (e.g., \texttt{Z.add} vs \texttt{BinInt.Z.add})
    \item \textbf{Implicit arguments}: Surface syntax may omit type parameters that are crucial for identification
    \item \textbf{Notations}: Operators like \texttt{+} can refer to different functions depending on type context
\end{itemize}

To address these challenges, we introduce a domain-specific tokenization system that assigns unique identifiers to Coq entities, ensuring semantic consistency across all naming variations:

\begin{definition}[Domain-specific Token]
\label{def:domain_token}
A token is a unique identifier $\text{id} \in \mathbb{N}$ assigned to each Coq entity. The tokenization function $\mathcal{T}$ $: \text{Names} \times \text{Context} \to \mathbb{N}$ maps an entity reference in a given context to its unique identifier, where:
\begin{itemize}
    \item $\text{Names}$ is the set of all possible names used to reference entities in Coq code
    \item $\text{Context}$ represents the resolution context (current module, section parameters, type environment)
    \item $\text{Token}$ satisfies semantic consistency: 
    \begin{itemize}
        \item If two names refer to the same Coq entity (even with different surface representations), they map to the same token
        \item If two names refer to different Coq entities, they map to different tokens
    \end{itemize}
\end{itemize}
\end{definition}

This tokenization enables reliable entity tracking across the entire Coq ecosystem. Whether analyzing dependencies in existing proofs, suggesting relevant lemmas during proof search, or ensuring correct imports, the system provides a consistent way to identify and reference Coq entities regardless of their surface representation.

These formal definitions provide the mathematical foundation for our data extraction and processing pipeline, ensuring precise representation of Coq's semantics throughout our framework.

\section{Evaluating Clarity}
\label{sec:evaluating_clarity}

To address our first question—\textit{Can we assess how well a model understands a given task?}—we develop a methodology for directly measuring how well models comprehend Coq concepts in theorem proving contexts. When a model fails to prove a theorem, it may be due to insufficient reasoning ability or inadequate clarity of the mathematical concepts involved. Our approach provides a systematic way to distinguish between these two possibilities.




\subsection{Experimental Design}

Given a proof scenario with various concepts, we:
\begin{enumerate}
    \item Extract all concepts appearing in the proof state
    \item Randomly select one concept
    \item Ask the model to provide its strict Coq definition
    \item Evaluate the semantic correctness of the generated definition
\end{enumerate}

For example, when evaluating clarity in the context of structured prompts(which include concepts like \texttt{nat}, \texttt{plus}, and \texttt{eq}), we might ask:
\begin{quote}
\textit{"Given the following structural proof context: [full structural prompt], please provide the strict Coq definition of the concept \texttt{plus}."}
\end{quote}

\subsection{Clarity Score}

We introduce the \textbf{Clarity Score}, a metric that quantifies how accurately a model can define concepts. Using a language model as an evaluator~\citep{zhang2024autonomous}, we assess:
\begin{quote}
\textit{"Is the following definition semantically correct for [concept]? [model's generated definition]"}\\
\textit{Answer: YES or NO}
\end{quote}

The Clarity Score is computed using log probabilities:
\begin{align}
\text{Clarity Score} = \frac{\exp(\log P(\text{YES}))}{\exp(\log P(\text{YES})) + \exp(\log P(\text{NO}))}
\end{align}

This produces a score between 0 and 1, where higher scores indicate better conceptual clarity. We implement this evaluation using the \texttt{DeepSeek-V3} model as the judge.

Our experiments (Section~\ref{sec:experiments}) demonstrate that current approaches leave models with limited comprehension of the mathematical concepts they encounter. This suggests that many reasoning failures may stem from inadequate task understanding rather than flawed logical reasoning.

This observation motivates our approach: by systematically enhancing how we present tasks to models, we may achieve better reasoning performance. The following section explores how to bridge this understanding gap through enriched task descriptions.
\section{Enhancing Task Description}
\label{sec:enhancing_task_description}

We now address the second question: \textit{Can we improve the model's understanding by enriching the task description?} Our key insight is that while human readers leverage implicit mathematical knowledge to interpret Coq code, language models require explicit access to the same semantic information that Coq's type checker uses internally.

\subsection{The Semantic Gap Problem}

Current approaches to LLM-based theorem proving face a fundamental limitation: they train models on surface-level \texttt{.v} files, missing the rich type-theoretic information that gives Coq code its precise meaning. Consider a simple expression like \texttt{a + b}:

\begin{itemize}
    \item \textbf{Surface syntax hides crucial information}: A term like \texttt{a + b} could represent natural number addition, list concatenation, or boolean operations depending on context
    \item \textbf{Implicit arguments remain hidden}: Coq infers many type parameters that are essential for clarity but invisible in source code
    \item \textbf{Module aliasing creates ambiguity}: The same entity can be referenced through multiple names (e.g., \texttt{Z.add} vs \texttt{BinInt.Z.add})
    \item \textbf{Notations obscure underlying structure}: Operators like \texttt{+} are syntactic sugar that map to different functions based on type context
\end{itemize}

These issues compound in real proofs, where clarity requires resolving types, tracking implicit arguments, and disambiguating overloaded notations—precisely the computations Coq performs internally but never exposes in source files.

\subsection{Coq Data Processing Pipeline}
\label{subsec:coq_pipeline}

To bridge this semantic gap, we develop a pipeline that intercepts Coq's compilation process to extract the type-theoretic information computed internally. Our approach transforms raw Coq source into structured representations that make implicit knowledge explicit, providing language models with the same semantic precision available to Coq's kernel.

The pipeline addresses three complementary aspects of semantic clarity:

\begin{enumerate}
    \item \textbf{Entity Extraction}: Extracts Coq entities as defined in Definition~\ref{def:coq_entity}, capturing the complete tuple $(n, c, o, i)$ including both surface representations and kernel-elaborated types.
    \item \textbf{Proof State Extraction}: Extracts Proofs as defined in Definition~\ref{def:interactive_proof}, recording complete proof state transitions with both surface and internal representations of goals and hypotheses.
    \item \textbf{Domain-specific Tokenizer}: Realizes the tokenization function $\mathcal{T}$ from Definition~\ref{def:domain_token}, ensuring semantic consistency across naming variations and context-dependent references.
\end{enumerate}

\textbf{How Does This Improve Clarity?} This structured context directly addresses each component of the semantic gap:

\begin{itemize}
    \item \textbf{Notation Clarity}: Models receive explicit mappings from surface notation to underlying functions, eliminating guesswork about operator meanings
    \item \textbf{Type Precision}: Complete type information enables models to reason about type compatibility and implicit argument instantiation
    \item \textbf{Reference Resolution}: Unique identifiers for entities prevent confusion from module aliasing and namespace conflicts
    \item \textbf{Proof Transparency}: Access to internal proof state representations reveals the complete logical context for each tactic decision
\end{itemize}

By providing this structured semantic context, we transform the theorem proving task from parsing ambiguous surface syntax to reasoning with precise mathematical semantics. The complete technical implementation details are provided in Appendix~\ref{sec:appendix_technical}.

\subsection{From Extracted Data to Structured Reasoning}
The semantic context extracted by our pipeline provides the foundation for enhanced task descriptions, but raw semantic information alone is insufficient. We must transform this data into structured prompts that guide models toward systematic mathematical reasoning. Our approach mirrors how expert Coq users mentally process proofs: first clarify the types and definitions involved, then track how tactics transform the proof state, and finally select appropriate next steps based on this understanding.

Our enhanced task descriptions follow a systematic format:

\begin{enumerate}
    \item \textbf{Proof State}: Current goals and hypotheses with dual representations
    \item \textbf{Entity Definitions}: For each referenced concept(examples in Appendix~\ref{sec:appendix_examples}):
        \begin{itemize}
            \item Origin: Source code definition
            \item Internal: Kernel representation
            \item Intuition: Natural language explanation
        \end{itemize}
    \item \textbf{Context Information}: Module imports, notation definitions, and available theorems
    \item \textbf{Additional Enhancements}: Other task-specific enhancements detailed in subsequent section~\ref{sec:searching_algorithm}
\end{enumerate}

These components are integrated into structured prompts that provide comprehensive semantic context for theorem proving tasks. The detailed format and examples of our structured prompts are provided in Appendix~\ref{sec:appendix_prompt_format}.

With both the evaluation methodology for understanding (Section~\ref{sec:evaluating_clarity}) and the enhancement pipeline now in place, we are ready to develop proof search algorithms that leverage these structured representations. The next section presents our Planner-Executor architecture that transforms this enhanced understanding into effective theorem proving capability. The complete experimental validation of our approach—demonstrating that enhanced task descriptions indeed improve both understanding scores and theorem proving performance—is presented in Section~\ref{sec:experiments}.
\section{Searching Algorithm}
\label{sec:searching_algorithm}

With enhanced clarity through structured semantic information, we now address how to leverage this clarity for effective theorem proving. We introduce a specialized Planner-Executor architecture that transforms our structured semantic data into systematic proof construction, explicitly modeling the hierarchical nature of mathematical reasoning.

\subsection{Planner-Executor Proof Architecture}
\label{subsec:planner_executor}

The architecture operates on two fundamental principles:

\begin{enumerate}
    \item \textbf{Structured Data Foundation}: Both Planner and Executor receive the rich structured representation:
    \begin{itemize}
        \item Current proof state $\sigma$ with dual surface/internal representations (Definition~\ref{def:proof_state})
        \item All entities referenced in the proof state and current context, with complete definitions $(n, c, o, i)$ retrieved via our semantic tokenizer (Definition~\ref{def:coq_entity})
    \end{itemize}
    
    \item \textbf{Hierarchical Decomposition}: The Planner analyzes this structured data to generate strategic guidance, which the Executor then uses—alongside the same structured data—to produce concrete tactics.
    \begin{itemize}
        \item \textbf{Planner Output}: Structured analysis including:
        \begin{itemize}
            \item Core mathematical concepts and structures involved
            \item Relevant theorems and properties applicable to the current state
            \item Proof techniques suitable for the goal (induction, contradiction, case analysis)
            \item Key relationships between hypotheses and goals
            \item Strategic summary synthesizing the analysis
        \end{itemize}
        \item \textbf{Executor Output}: Multiple concrete Coq tactics generated via beam search
    \end{itemize}
\end{enumerate}

This hierarchical decomposition transforms our enhanced clarity into actionable proof strategies. While the semantic extraction provides the "what" (precise mathematical meanings), the Planner-Executor architecture provides the "how" (systematic proof construction).

\subsection{Enhanced Capabilities}

Beyond the core Planner-Executor decomposition, additional techniques further amplify the architecture's effectiveness (detailed algorithms provided in Algorithm~\ref{alg:proof_search_part1}):

\textbf{Search and Exploration:}
\begin{itemize}
    \item \textbf{Beam Search}: Maintains the $k$ most promising proof states (width=3), efficiently exploring multiple proof paths while pruning less convincing branches
    \item \textbf{Error Reflection}: Compiler error messages from failed tactics guide regeneration, enabling iterative refinement based on concrete feedback
    \item \textbf{Precise External Referencing}: Our tokenizer enables us to trace any theorem back to its source file and automatically generate the correct \texttt{Require} statements when referencing external premises

\end{itemize}

\textbf{Context Augmentation:}
\begin{itemize}
    \item \textbf{Retrieval Augmentation}: Encodes the Planner's strategic analysis to retrieve semantically similar premises and tactics from the corpus
    \item \textbf{Precise External Referencing}: Our semantic tokenizer ensures accurate resolution of all entity references, preventing naming ambiguities across different namespaces
    \item \textbf{Intuitive Annotations}: Each entity is augmented with single-sentence natural language descriptions for enhanced comprehension
\end{itemize}

\textbf{Proof Intelligence:}
\begin{itemize}
    \item \textbf{Dynamic Summarization}: Generates proof progress summaries including expected completion steps and state quality scores
    \item \textbf{Tactic Explanation}: Produces bilingual (formal/natural) explanations for each successful tactic application, capturing transformations and rationale
    \item \textbf{Tactic Trace}: Maintains complete proof paths with step-by-step explanations of how the current state was reached
    \item \textbf{Public Notebook}: Evaluates and maintains a fixed-size cache (15 items) of key insights from successful proof steps, accessible throughout the proof
\end{itemize}

These capabilities work together to create a comprehensive reasoning framework. The semantic foundation enables precise clarity, the Planner provides strategic direction, and the enhanced capabilities ensure robust execution even in the face of failures.

\subsection{Proof Search Algorithm}
\label{subsec:proof_search}

Having established the Planner-Executor architecture and its enhanced capabilities, we now present the complete proof search algorithm that operationalizes these components into a systematic theorem proving process.

\paragraph{Core Algorithm}
Our proof search integrates all previous components into a coherent search strategy. The algorithm maintains a beam of $B=3$ candidate proof states, using the structured semantic context to guide exploration at each step. 

Each search iteration performs three coordinated phases:

\begin{enumerate}
    \item \textbf{State Expansion}: For each candidate state in the beam:
    \begin{itemize}
        \item Extract structured representation: proof state $\sigma$, expanded entities via semantic tokens
        \item Generate strategic analysis using the Planner module
        \item Retrieve relevant premises and successful tactics based on the strategy
        \item Produce multiple tactic candidates using the Executor with beam search
        \item Validate tactics through Coq's compiler, applying error reflection on failures
    \end{itemize}
    
    \item \textbf{State Selection}: Score and rank all generated states, retaining top-$B$ based on:
    \begin{itemize}
        \item Progress toward goal completion
        \item Strategy coherence from Planner analysis
        \item Historical success patterns
    \end{itemize}
    
    \item \textbf{Context Maintenance}: Update shared resources:
    \begin{itemize}
        \item Public notebook with proven insights (max 15 items)
        \item Tactic trace for successful paths
        \item Error history for failed attempts
    \end{itemize}
\end{enumerate}

The algorithm terminates when any branch achieves $\sigma_k$ where all goals are discharged, or when reaching the depth limit. Detailed pseudocode is provided in Algorithm~\ref{alg:proof_search_part1}.

\begin{algorithm}[H]
\caption{Neural Theorem Proving with Structured Coq Representations (Part 1)}
\label{alg:proof_search_part1}
\begin{algorithmic}[1]
\Require Theorem $T$, Max depth $D$, Beam width $B$, Max retries $R$
\Ensure Proof trace or FAILURE

\State $S_0 \gets$ \Call{CompileTheorem}{$T$}
\State $\mathit{layer} \gets \{(S_0, [], \text{``''}, \text{``''})\}$ \Comment{(state, trace, summary, notes)}

\For{$d = 1$ \textbf{to} $D$}
    \State $\mathit{next} \gets \emptyset$
    \State $\mathit{insights} \gets []$
    
    \For{\textbf{each} $(S, \mathit{trace}, \mathit{summary}, \mathit{notes}) \in \mathit{layer}$}
        
        \State \textbf{/* Step 1: Extract concepts */}
        \State $C \gets$ \Call{ExtractConcepts}{$S, \mathit{depth}=1$}
        
        \State \textbf{/* Step 2: Generate strategy */}
        \State $\mathit{prompt\_method} \gets (S, C, \mathit{trace}, \mathit{summary}, \mathit{notes})$
        \State $(\mathit{resp}, \mathit{strategy}) \gets$ \Call{GenerateStrategy}{$\mathit{prompt\_method}$}
        
        \State \textbf{/* Step 3: Retrieve relevant information */}
        \State $(\mathit{premises}, \mathit{tactics}) \gets$ \Call{Retrieve}{$\mathit{resp}, k$}
        
        \State \textbf{/* Step 4: Generate candidate tactics */}
        \State $\mathit{prompt\_tactic} \gets (S, C, \mathit{premises}, \mathit{tactics}, \mathit{strategy}, \mathit{trace}, \mathit{summary}, \mathit{notes})$
        \State $\mathit{candidates} \gets$ \Call{BeamSearchTactics}{$\mathit{prompt\_tactic}, 10$}
        
        \State \textbf{/* Step 5: Validate and retry on failure */}
        \State $\mathit{valid} \gets []$
        \State $\mathit{failed} \gets []$
        
        \For{\textbf{each} $t \in \mathit{candidates}$}
            \State $r \gets$ \Call{CompileTactic}{$t, S$}
            \If{$r.\mathit{success}$}
                \State $\mathit{valid}.\text{append}(t)$
            \Else
                \State $\mathit{failed}.\text{append}((t, r.\mathit{error}))$
            \EndIf
        \EndFor
        \State \textbf{/* Retry failed tactics */}
        \State $\mathit{retry} \gets 0$
        \While{$\mathit{retry} < R$ \textbf{and} $\mathit{failed} \neq []$ \textbf{and} $|valid| \leq 10$}
            \State $\mathit{errors} \gets$ [errors from $\mathit{failed}$]
            \State $\mathit{prompt\_error} \gets (S, C, \mathit{errors}, \mathit{trace}, \mathit{prompt\_method})$
            \State $(\mathit{resp'}, \mathit{strategy'}) \gets$ \Call{GenerateStrategy}{$\mathit{prompt\_error}$}
            \State $(\mathit{premises'}, \mathit{tactics'}) \gets$ \Call{Retrieve}{$\mathit{resp'}, k$}
            \State $\mathit{prompt\_tactic'} \gets (S, C, \mathit{premises'}, \mathit{tactics'}, \mathit{strategy'}, \mathit{trace}, \mathit{summary}, \mathit{notes})$
            \State $\mathit{retry\_candidates} \gets$ \Call{BeamSearchTactics}{$\mathit{prompt\_tactic'}, 10$}
            \State $\mathit{failed} \gets []$
        \algstore{myalg} 
\end{algorithmic}
\end{algorithm}

\addtocounter{algorithm}{-1}
\begin{algorithm}[H]
\caption{Neural Theorem Proving with Structured Coq Representations (Part 2)}
\label{alg:proof_search_part2}
\begin{algorithmic}[1]
\algrestore{myalg} 
            \For{\textbf{each} $t \in \mathit{retry\_candidates}$}
                \State $r \gets$ \Call{CompileTactic}{$t, S$}
                \If{$r.\mathit{success}$}
                    \State $\mathit{valid}.\text{append}(t)$
                \Else
                    \State $\mathit{failed}.\text{append}((t, r.\mathit{error}))$
                \EndIf
            \EndFor
            \State $\mathit{retry} \gets \mathit{retry} + 1$
        \EndWhile
        \State \textbf{/* Step 6: Apply valid tactics */}
        \For{\textbf{each} $t \in \mathit{valid}$}
            \State $S' \gets$ \Call{ApplyTactic}{$t, S$}
            
            \If{\Call{IsGoalComplete}{$S'$}}
                \State \Return $\mathit{trace} + [t]$
            \EndIf
            
            \If{\Call{IsSubgoalComplete}{$S'$}}
                \State $S' \gets$ \Call{RefreshWithIdtac}{$S'$}
            \EndIf
            
            \State $\mathit{expl} \gets$ \Call{GenerateExplanation}{$S, t, S'$}
            \State $\mathit{trace'} \gets \mathit{trace} + [(t, \mathit{expl})]$
            \State $\mathit{summary'} \gets$ \Call{SummarizeProof}{$\mathit{trace'}$}
            
            \State $\mathit{next}.\text{append}((S', \mathit{trace'}, \mathit{summary'}, \mathit{notes}))$
            \State $\mathit{insights}.\text{append}(\mathit{expl})$
        \EndFor
    \EndFor
    
    \If{$\mathit{next} = \emptyset$}
        \State \Return FAILURE
    \EndIf
    
    \State \textbf{/* Step 7: Update public notes */}
    \If{$\mathit{insights} \neq []$}
        \State $\mathit{notes} \gets$ \Call{UpdatePublicNotes}{$S_0, \mathit{insights}, \mathit{notes}$}
    \EndIf
    
    \State \textbf{/* Step 8: Beam search selection */}
    \If{$|\mathit{next}| > B$}
        \State $\mathit{layer} \gets$ \Call{SelectBest}{$S_0, \mathit{next}, B$}
    \Else
        \State $\mathit{layer} \gets \mathit{next}$
    \EndIf
\EndFor

\State \Return FAILURE

\end{algorithmic}
\end{algorithm}

The algorithm operationalizes the insights from our enhanced clarity: it uses the dual representations to guide search, leverages the Planner's analysis for strategic coherence, and employs beam search to balance exploration with computational efficiency.

\subsection{Integrating Components into a Complete Proof System}

This section completes our methodological framework by integrating all previously developed components into a unified theorem proving system. We have now established:

\textbf{Semantic Foundation}: Our data processing pipeline (Section~\ref{subsec:coq_pipeline}) extracts the precise type-theoretic information that Coq computes internally, providing models with disambiguated entities, complete type information, and resolved proof states.

\textbf{Strategic Architecture}: Our Planner-Executor framework (Section~\ref{subsec:planner_executor}) separates high-level proof strategy from low-level tactic generation, enabling systematic reasoning that mirrors expert mathematical practice.

\textbf{Coordinated Search}: Our proof search algorithm orchestrates these components into a coherent search process, where semantic clarity guides strategic planning, which in turn directs tactical execution.

The resulting system represents a complete methodology for neural theorem proving that leverages structured semantic information at every level—from individual symbol disambiguation to high-level proof strategy. With our approach fully specified, we are now ready to evaluate its effectiveness. The next section presents comprehensive experiments that measure both clarity improvement and theorem proving performance, providing empirical answers to our three central research questions.
\section{Experiments}
\label{sec:experiments}

Having established our complete methodological framework, we now conduct comprehensive experiments to validate our three research questions:
\begin{enumerate}
    \item \textbf{Can we assess how well a model understands a given task?} We evaluate our clarity measurement methodology (Section~\ref{sec:evaluating_clarity}) across different information configurations.
    \item \textbf{Can we improve the model's understanding by enriching the task description?} We test whether our structured semantic information (Section~\ref{sec:enhancing_task_description}) enhances both clarity and reasoning performance.
    \item \textbf{Does deeper clarity lead to better reasoning performance?} We analyze the relationship between clarity scores and theorem proving success rates, validating our central hypothesis.
\end{enumerate}

Our experimental design systematically addresses each question in sequence, building evidence for our approach's effectiveness.

\subsection{Experiment 1: Assessing Model Clarity and Task Enhancement}
\label{sec:exp_clarity_assessment}

Our first experiment simultaneously addresses Questions 1 and 2 by measuring model clarity across different information configurations. This experiment validates our measurement methodology while demonstrating the effectiveness of task enrichment.

\subsubsection{Experimental Setup}

For evaluation, we construct a dataset from 1,000 structured prompts randomly sampled from our Coq proof generation process. Each structured prompt contains the components described in Section~\ref{sec:enhancing_task_description}, including a list of global definitions referenced in the proof context. We use this reference list to identify relevant semantic content, randomly selecting up to 3 definitions per prompt (or all if fewer than 3) to construct the corresponding information configurations. Consequently, each information condition consists of slightly fewer than 3,000 evaluated instances. The detailed format of our structured prompts is provided in Appendix~\ref{sec:appendix_prompt_format}.

\subsubsection{Information Configurations}

We test three types of semantic information extracted by our pipeline:

\begin{itemize}
    \item \textbf{Origin}: Raw Coq source code definitions
    \item \textbf{Internal}: Compiler internal representations (kernel-level)
    \item \textbf{Intuition}: Natural language explanations generated by large language models
\end{itemize}

This yields 8 different information combinations, plus additional control conditions:

\begin{enumerate}
    \item \textbf{No Context}(current standard approach): Only proof state with simple names (e.g., \texttt{plus})
    \item \textbf{Qualified Name}: Only proof state with full names (e.g., Coq.Arith.Plus.plus)
    \item \textbf{Empty Reference}: Proof state + additional techniques
    \item \textbf{Origin Only}: Proof state + raw Coq definitions + additional techniques
    \item \textbf{Internal Only}: Proof state + kernel representations + additional techniques
    \item \textbf{Intuition Only}: Proof state + natural language explanations + additional techniques
    \item \textbf{Origin + Internal}: Combination of raw code and kernel info + additional techniques
    \item \textbf{Origin + Intuition}: Combination of raw code and explanations + additional techniques
    \item \textbf{Internal + Intuition}: Combination of kernel info and explanations + additional techniques
    \item \textbf{Complete Information}: All three types of information + additional techniques
    \item \textbf{Chinese Translation}: Complete information translated to Chinese + additional techniques
\end{enumerate}

Before presenting our enhanced configurations, it is crucial to establish the baseline. Current state-of-the-art theorem proving systems, including DeepSeek Prover~\citep{xin2024deepseek}, Kimina-Prover~\citep{kimina_prover_2025}, and other advanced neural theorem provers~\citep{wang2023lego,wu2024internlm2}, employ a standard approach where models are provided with \textbf{No Context}.

Our evaluation reveals that under these standard conditions—the same setup used by current theorem proving approaches—models achieve only 44.5\% understanding when asked to define basic concepts. This means that nearly 60\% of the time, models cannot accurately comprehend the mathematical concepts they are working with, which may explain why reasoning performance often falls short of expectations.


\subsubsection{Clarity Measurement Results}

Using our concept definition task methodology, we measure clarity scores across all configurations. The results demonstrate that our measurement approach effectively captures differences in model comprehension:

\begin{table}[htbp]
    \centering
    \begin{tabular}{lc}
        \toprule
        \textbf{Information Configuration} & \textbf{Clarity Score} \\
        \midrule
        \multicolumn{2}{l}{\textit{Control Conditions}} \\
        No Context(current standard approach) & 0.445 \\
        Qualified names & 0.581 \\
        \midrule
        \multicolumn{2}{l}{\textit{Empty Information}} \\
        Empty Reference & 0.615 \\
        \midrule
        \multicolumn{2}{l}{\textit{Single Information Type}} \\
        Origin Only & 0.798 \\
        Internal Only & 0.712 \\
        Intuition Only & 0.667 \\
        \midrule
        \multicolumn{2}{l}{\textit{Combined Information}} \\
        Origin + Internal & 0.815 \\
        Origin + Intuition & 0.803 \\
        Internal + Intuition & 0.714 \\
        Complete Information & \textbf{0.823} \\
        \midrule
        \multicolumn{2}{l}{\textit{Language Variation}} \\
        Chinese Translation & 0.732 \\
        \bottomrule
    \end{tabular}
    \caption{Model clarity scores under different information configurations}
    \label{tab:clarity_scores}
\end{table}

The results validate our clarity measurement methodology and reveal several important insights:

\begin{enumerate}
    \item \textbf{Measurement sensitivity}: Our methodology successfully discriminates between different levels of clarity, with scores ranging from 44.5\% to 82.3\%.
    
    \item \textbf{Qualified names matter}: Using fully qualified names instead of simple names improves clarity by 13.6\% absolute, demonstrating that disambiguation is crucial.
    
    \item \textbf{Structured information effectiveness}: Providing origin code definitions improves clarity from 61.5\% to 79.8\%—an 18.8\% absolute gain, showing our semantic extraction provides valuable context.
    
    \item \textbf{Cross-language validity}: Chinese translations maintain substantial clarity capability (73.2\%), indicating that structural information rather than language-specific patterns drives improvements.
    
    \item \textbf{Information complementarity}: While origin code provides the strongest single signal, internal representations and natural language intuitions offer complementary benefits, with complete information achieving 82.3\% clarity.
\end{enumerate}

These findings address Questions 1 and 2 directly: our methodology can effectively assess model understanding across different configurations, and enriching task descriptions with structured semantic information dramatically improves comprehension from 44.5\% to 82.3\%.

\subsection{Experiment 2: Clarity-Performance Relationship}
\label{sec:main_results}

\subsubsection{Experimental Setup}

\paragraph{Dataset} Following Graph2Tac's evaluation protocol, we randomly sample 10\% from their test set. Due to differences in data processing pipelines, the final dataset contains 1,300 theorems from 15 standard Coq packages, which we believe is sufficient for meaningful comparison. The dataset covers diverse mathematical domains including:
\begin{itemize}
    \item Pure mathematics: real analysis, number theory, topology, group theory
    \item Computer science: separation logic, type theory, formal verification
    \item Computational mathematics: constructive algebra, homotopy type theory
\end{itemize}

\paragraph{Search Parameters} Based on preliminary experiments, we set:
\begin{itemize}
    \item Maximum proof depth $D = 15$ (sufficient for most theorems)
    \item Beam width $B = 3$ (balancing exploration and efficiency)
    \item Maximum retries $R = 3$ (one retry with error feedback)
\end{itemize}

\paragraph{Computational Budget}
To ensure fair comparison across different model configurations, we establish a consistent computational budget for proof search. Each theorem is allocated a maximum of 860 tactic evaluations, calculated as follows:

\begin{align}
\text{Budget} &= \text{Initial layer} \times \text{Tactics per state} \times \text{Reconsider factor} \\
&\quad + \text{Subsequent layers} \times \text{Beam width} \times \text{Tactics per state} \times \text{Reconsider factor} \\
&= 1 \times 10 \times 2 + 14 \times 3 \times 10 \times 2 = 860
\end{align}


\subsubsection{Theorem Proving Performance}

We evaluate our complete system against the Graph2Tac baseline to measure the practical impact of enhanced clarity:

\begin{table}[htbp]
    \centering
    \begin{tabular}{lcccc}
        \toprule
        \textbf{Method} & \textbf{Success Rate} & \textbf{Avg Depth} & \textbf{Avg Tactics} \\
        \midrule
        Graph2Tac (baseline) & 33.2\% & - & - \\
        DeepSeek-V3 (our baseline) & 21.8\% & 3.2 & 485 \\
        DeepSeek-V3 + DeepSeek-V3 (our method) & \textbf{45.8\%} & 2.7 & 357 \\
        \bottomrule
    \end{tabular}
    \caption{Main results comparing our method to Graph2Tac baseline}
    \label{tab:main_comparison}
\end{table}

\begin{itemize}
    \item \textbf{Significant improvement over state-of-the-art}: Our Planner-Executor system achieves 45.8\% success rate, outperforming Graph2Tac by 12.6 percentage points (38\% relative improvement)
    \item \textbf{Efficiency gains}: Our method requires fewer tactics on average (357 vs 485 for baseline) and achieves proofs in fewer steps (2.7 vs 3.2 depth)
    \item \textbf{Bridging the gap}: While our DeepSeek-V3 baseline (21.8\%) initially underperforms Graph2Tac, our structured semantic approach not only closes this gap but achieves superior performance
\end{itemize}

\subsubsection{Clarity-Performance Correlation}
\label{sec:correlation_analysis}

To directly test our central hypothesis, we analyze the relationship between understanding scores and theorem proving success rates across different information configurations. For this analysis, we use a randomly sampled subset of 100 theorems from our full dataset of 1,300 theorems to ensure computational feasibility while maintaining statistical validity:

\begin{table}[htbp]
    \centering
    \begin{tabular}{lcc}
        \toprule
        \textbf{Configuration} & \textbf{Clarity Score} & \textbf{Success Rate} \\
        \midrule
        No Context(current standard approach) & 0.445 & 21\% \\
        Qualified Name & 0.581 & 25\% \\
        Internal Only & 0.712 & 38\% \\
        Origin Only & 0.798 & 42\% \\
        Complete Information & 0.823 & 45\% \\
        \bottomrule
    \end{tabular}
    \caption{Strong correlation between clarity and proving performance}
    \label{tab:clarity_correlation}
\end{table}

The correlation analysis reveals a remarkably strong linear relationship (correlation coefficient $r = 0.98$) between understanding and reasoning performance. As understanding scores increase from 44.5\% to 82.3\%, theorem proving success rates improve proportionally from 21\% to 45\%.

Combined with our performance results—where enhanced understanding achieves 45.8\% success rate versus the 33.2\% Graph2Tac baseline—this evidence addresses Question 3 conclusively. The near-perfect correlation demonstrates that understanding improvements directly drive reasoning performance, supporting our central hypothesis that AI reasoning failures in formal mathematics often stem from inadequate task understanding rather than insufficient reasoning capability.

\subsection{Additional Results and Analysis}
\label{sec:finetune_results}

Having answered our three central research questions, we present additional experiments that demonstrate the broader applicability and practical value of our approach.

\subsubsection{Fine-Tuned Model Performance}

We investigate whether fine-tuning smaller models on our structured data can achieve competitive performance, offering efficiency benefits for practical deployment.

\subsubsection{Training Data}
Our fine-tuning dataset consists of 400,000 (\textit{structured\_context}, \textit{tactic}) pairs extracted from our data processing pipeline, where:
\begin{itemize}
    \item \textit{structured\_context} includes all components: current proof state with dual representations, referenced entities with complete definitions, strategic analysis, retrieved premises/tactics, proof trace, and error history  
    \item \textit{tactic} is the successful tactic applied at that state
\end{itemize}

\subsubsection{Results}
The fine-tuning experiment yields promising results:

\begin{table}[htbp]
    \centering
    \begin{tabular}{lcccc}
        \toprule
        \textbf{Method} & \textbf{Success Rate} & \textbf{vs Graph2Tac} & \textbf{Model Size} & \textbf{Efficiency} \\
        \midrule
        Graph2Tac (baseline) & 33.2\% & 1.00× & - & - \\
        DeepSeek-V3 + DeepSeek-V3 & 45.8\% & 1.38× & 671B & 1× \\
        Qwen-2.5-7B (fine-tuned) & 45.2\% & 1.36× & 7B & 96× \\
        Qwen-2.5-32B (fine-tuned) & \textbf{48.6\%} & \textbf{1.46×} & 32B & 21× \\
        \bottomrule
    \end{tabular}
    \caption{Fine-tuned models achieve competitive performance with significant parameter efficiency}
    \label{tab:finetune_results}
\end{table}

Despite using only 400,000 training pairs—a relatively small dataset—the fine-tuning results are promising. The Qwen-2.5-32B model achieves 48.6\% success rate, our best overall result, while the 7B model nearly matches the 671B DeepSeek-V3 performance with 96× fewer parameters. This demonstrates the efficiency potential of our structured data approach.

Given that our training data represents only a fraction of available proof data, there is substantial room for improvement. With more comprehensive training data, we anticipate even stronger performance from fine-tuned models, making this an attractive direction for efficient deployment.

\subsubsection{Detailed Analysis by Library Type}

Our results show consistent improvements across different mathematical domains:

\begin{table}[htbp]
    \centering
    \begin{tabular}{lccc}
        \toprule
        \textbf{Library} & \textbf{DS Baseline} & \textbf{DS+DS} & \textbf{Qwen-32B-FT} \\
        \midrule
        \textbf{Logic \& Foundations} & & & \\
        TLC & 0.40 & 0.65 & 0.68 \\
        SMTCoq & 0.31 & 0.55 & 0.65 \\
        HoTT & 0.19 & 0.39 & 0.40 \\
        \midrule
        \textbf{Pure Mathematics} & & & \\
        CoRN & 0.35 & 0.52 & 0.63 \\
        Topology & 0.10 & 0.27 & 0.26 \\
        Ceres & 0.17 & 0.29 & 0.57 \\
        \midrule
        \textbf{Computer Science} & & & \\
        PolTac & 0.64 & 0.81 & 0.93 \\
        Qcert & 0.23 & 0.57 & 0.65 \\
        iris & 0.10 & 0.09 & 0.20 \\
        \bottomrule
    \end{tabular}
    \caption{Success rates by library demonstrating broad applicability}
    \label{tab:library_breakdown}
\end{table}

Key observations:
\begin{itemize}
    \item \textbf{Consistent improvements}: Both DS+DS and fine-tuned approaches show gains across all library types
    \item \textbf{Domain specialization}: Fine-tuned models excel particularly in specialized domains like PolTac (tactics) and Ceres (geometry)
    \item \textbf{Challenging domains}: Even in difficult areas like topology and HoTT, our approach provides meaningful improvements
\end{itemize}

\subsection{Ablation Studies} 
\label{sec:ablation_studies}

We conducted comprehensive ablation studies to understand the contribution of each component. The information component analysis is performed on 1386 theorems, which are randomly sampled 10\% from the Coq Standard Library. Others are performed on the Ceres library containing 78 theorems.

\paragraph{Information Component Analysis}

\begin{table}[htbp]
    \centering
    \begin{tabular}{lc}
        \toprule
        \textbf{Information Components} & \textbf{Success Rate} \\
        \midrule
        Proof State + Entities + Structured Internal Representations & 37.2\% \\
        + Retrieved Premises and Tactics & 42.6\% (+5.4\%) \\
        + Proof Trace & 46.8\% (+4.2\%) \\
        + Public Notebook & \textbf{48.6\%} (+1.8\%) \\
        \bottomrule
    \end{tabular}
    \caption{Incremental benefits of information components}
    \label{tab:information_ablation}
\end{table}

Each information component provides meaningful improvements:
\begin{itemize}
    \item \textbf{Retrieved premises/tactics} (+5.4\%): Largest single improvement from relevant context
    \item \textbf{Proof trace} (+4.2\%): Historical context aids strategic planning
    \item \textbf{Public notebook} (+1.8\%): Shared insights across proof steps
\end{itemize}

\paragraph{Architecture Comparison}

\begin{table}[htbp]
    \centering
    \begin{tabular}{lc}
        \toprule
        \textbf{Architecture} & \textbf{Success Rate} \\
        \midrule
        Single DeepSeek-V3 & 21.8\% \\
        DeepSeek-V3 + DeepSeek-V3 (Planner-Executor) & 45.8\% \\
        DeepSeek-V3 (Planner) + Qwen-32B-FT (Executor) & \textbf{48.6\%} \\
        \bottomrule
    \end{tabular}
    \caption{Architectural design impact on performance(78 theorems from Ceres library)}
\end{table}

The Planner-Executor decomposition provides substantial benefits (+24\%), with fine-tuned executors offering additional gains (+2.8\%).

\paragraph{Search Strategy Optimization}

\begin{table}[htbp]
    \centering
    \begin{tabular}{lcc}
        \toprule
        \textbf{Retry Count} & \textbf{State Selection} & \textbf{Success Rate} \\
        \midrule
        0 & Shortest proof states & 35.1\% \\
        3 & Shortest proof states & 40.3\% \\
        1 & Model-based selection & 44.2\% \\
        \bottomrule
    \end{tabular}
    \caption{Search strategy optimization results(78 theorems from Ceres library)}
\end{table}

A single retry with error feedback provides optimal performance, while excessive retries hurt due to search space explosion.





\section{Conclusion}
This work investigates how well models understand mathematical tasks by introducing clarity score. Using this metric, we show that adding structured semantic context significantly improves clarity score. In addition, our experiments reveal a strong positive correlation between clarity score and reasoning ability. 
Notably, our method yields substantial gains for general-purpose language models like DeepSeek V3 --- demonstrating that these models benefit not just from scale, but from principled task representations. 
While developed for mathematics, our approach generalizes to other domains that demand precise reasoning, such as formal verification, clinical decision-making, and software engineering. Moreover, it is model-agnostic, offering a path toward more principled, structure-aware reasoning systems of the future.



\bibliographystyle{plain}  
\bibliography{references}   


\appendix
\section{Technical Implementation Details}
\label{sec:appendix_technical}
This part provides detailed technical information about our data processing pipeline and implementation choices.

\subsection{Entity Extraction}

\paragraph{Implementation Overview}
Coq's compilation pipeline transforms user-written definitions through multiple stages:
$$\text{Raw AST} \xrightarrow{\text{parsing}} \texttt{glob\_constr} \xrightarrow{\text{pre-typing}} \texttt{constr} \xrightarrow{\text{type inference}} \texttt{typed\_constr}$$

At the \texttt{typed\_constr} stage, Coq has performed complete type inference and elaboration—all implicit arguments are computed, universe levels are instantiated, and every subterm has been assigned its precise type. We intercept at this stage to capture the fully-elaborated internal representation ($i$) alongside the original surface syntax ($o$), while the entity type ($c$) and name ($n$) are extracted during vernacular command processing.

\paragraph{Kernel Term Processing}
Every high-level Coq construct (theorems, definitions, fixpoints, etc.) is internally compiled down to expressions built from 18 fundamental kernel constructors. For example, a theorem statement becomes a \texttt{Prod} (forall) expression, while its proof term might combine \texttt{Lambda} (function), \texttt{App} (application), and \texttt{Ind} (inductive type) constructors. We implement a recursive \texttt{constr\_to\_string} method that processes these kernel terms at three levels of detail:
\begin{itemize}
    \item \textbf{Pure constr}: Raw kernel representation
    \item \textbf{Standard constr}: Kernel terms with resolved variable names  
    \item \textbf{Processed}: Simplified representation for language models
\end{itemize}

\paragraph{Local Variable Context Extraction}
During the compilation process, Coq automatically infers and tracks the types of all local variables. When we intercept at the \texttt{glob\_constr} stage during pre-typing, we capture all \texttt{GVar} references along with their inferred types. This type information becomes naturally embedded in the internal representation.

This context extraction is crucial for semantic clarity. Consider a theorem stating $a + b = b + a$:
\begin{itemize}
    \item With $a, b : \mathbb{N}$, this is commutativity of natural number addition
    \item With $a, b : \text{list}\ A$, this would be about list concatenation (likely false)
    \item With $a, b : \text{bool}$, this might represent logical operations
\end{itemize}

The same syntactic pattern has entirely different meanings depending on types. By extracting type annotations implicit in surface syntax, we provide models with the semantic context necessary for type-aware reasoning.

\paragraph{Key Technical Decisions}

\begin{itemize}
    \item \textbf{Environment Trimming for De Bruijn Index Resolution}: In Coq's kernel representation, variables are represented using De Bruijn indices—a nameless representation where variables are identified by their position in the binding context rather than by name. For example, in the term $\lambda x. \lambda y. x + y$, the variable $y$ is represented by index 0 (most recently bound) and $x$ by index 1 (bound one level up).
    
    To correctly resolve these indices back to meaningful variable names, we must maintain a precise environment that mirrors the binding structure. Consider processing the term $\forall (x : \text{nat}), \forall (y : \text{nat}), x + y = y + x$:
    \begin{itemize}
        \item When processing the outer $\forall$, the environment is empty
        \item After binding $x$, the environment becomes $[x]$
        \item After binding $y$, the environment becomes $[y, x]$ (with $y$ at index 0)
        \item When encountering index 1 in the body, we resolve it to $x$
        \item When encountering index 0, we resolve it to $y$
    \end{itemize}
    
    The challenge arises with Coq's global environment containing thousands of definitions. When processing nested terms, we must trim this global context to match each subterm's local binding context. This trimming ensures that De Bruijn indices correctly map to their intended variables, handling the complexity of deeply nested scopes, pattern matching branches, and local definitions.

    \item \textbf{Kernel Name Disambiguation}: Coq's module system creates naming complexity—the same entity can have different user-facing names. For instance, \texttt{Z.quotrem} might be defined in \texttt{BinIntDef} but re-exported through \texttt{BinInt}. We create globally unique identifiers by combining the kernel name (internal unique identifier) with the canonical path (user-facing name):
    \begin{center}
    \texttt{Coq.ZArith.BinInt.Z.quotrem<ker>Coq.ZArith.BinIntDef.Z.quotrem}
    \end{center}
    This ensures unambiguous entity references regardless of import paths or module aliases.

\end{itemize}

Through this extraction pipeline, we capture the complete Coq Entity tuple $(n, c, o, i)$ from Definition~\ref{def:coq_entity}. By intercepting Coq's compilation at multiple stages and implementing comprehensive kernel term processing, we provide language models with parallel representations—surface syntax alongside fully-elaborated internal forms—enabling them to learn the relationship between what users write and what Coq actually computes.

\subsection{Proof State Extraction}

\paragraph{Implementation Overview}
To capture the interactive proof sequences defined in Definition~\ref{def:interactive_proof}, we instrument Coq's tactic interpreter to record proof states before and after each tactic application. For each tactic $T_i$, we capture the complete proof state $\sigma_{i-1}$ at the entry point and $\sigma_i$ after execution, extracting both surface and internal representations $(H^o, H^i, G^o, G^i)$ for every goal. The internal representations are processed using the same \texttt{constr\_to\_string} method from entity extraction, as hypotheses and goals are ultimately combinations of the same 18 kernel constructors.

\paragraph{Key Technical Decisions}

\begin{itemize}
    \item \textbf{Tactic Linearization}: Coq's semicolon operator allows parallel goal processing, where a single tactic can be applied to multiple goals simultaneously. This creates non-linear proof structures that are challenging for sequence models to learn. We modify Coq's interpreter to use 
    \texttt{Goal.enter}, which isolates each goal in its own environment and processes them sequentially. This linearization transforms complex parallel proof steps into sequences that align with Definition~\ref{def:interactive_proof}. For rare cases with sequential dependencies between goals (less than 1\% of proofs), we provide a compiler flag to revert to parallel processing.
    \begin{itemize}
        \item \textbf{Data Amplification}: This linearization significantly amplifies training data—a single compound tactic operating on $n$ goals in parallel now generates $n$ distinct proof state transitions, exposing intermediate states that would otherwise be hidden. This fine-grained decomposition provides models with detailed proof trajectories essential for learning tactical reasoning.
    \end{itemize}
    \item \textbf{Ltac Expansion}: User-defined Ltac tactics appear as single atomic steps in surface-level proofs but often expand into complex sequences of primitive tactics. We implement configurable-depth expansion of these tactics, recursively unfolding them to reveal their underlying proof patterns. This expansion is particularly important for tactics with branching structures, where we ensure consistent depth tracking across all branches to maintain structural clarity.
    
    \item \textbf{Meaningful Transition Filtering}: We validate state transitions by tracking changes in goal structures and existential variable (evar) maps between proof states. This ensures we capture all tactics that modify the proof state, including those that instantiate evars or restructure goals without visibly changing the goal count. Only tactics that genuinely transform the proof state are recorded, providing models with semantically meaningful proof steps.
\end{itemize}

This extraction pipeline captures complete interactive proofs as defined in Definition~\ref{def:interactive_proof}, with each proof step containing full before/after states including all hypotheses and goals in both surface and internal forms. The linearization and expansion mechanisms expose the fine-grained structure of proof construction, enabling language models to learn from detailed proof trajectories rather than opaque high-level scripts.

\subsection{Domain-specific Tokenizer}

\paragraph{Design Principle}
With entities and proof states extracted, we construct a tokenizer that preserves Coq's semantic identity. The core principle, formalized in Definition~\ref{def:domain_token}, ensures that tokens represent semantic meaning rather than surface syntax: identical Coq entities receive the same token regardless of how they are referenced (e.g., \texttt{Nat.add} and \texttt{Coq.Init.Nat.add}), while distinct entities always receive different tokens.

\paragraph{Token Categories}
Our tokenizer assigns unique integer identifiers to:
\begin{itemize}
    \item \textbf{Global Identifiers}: Fully-qualified names from extracted entities.
    \begin{itemize}
        \item Inductive types: Each constructor receives its own global ID (e.g., \texttt{nat}, \texttt{O}, and \texttt{S} are assigned distinct tokens)
        \item Recursive definitions: Individual IDs for each function in mutually recursive groups, preserving their semantic independence
        \item Namespace resolution: Pattern matching strips module and section prefixes to distinguish truly global entities from locally-scoped definitions that appear with qualified names due to Coq's scoping mechanisms
    \end{itemize}
    \item \textbf{Local Variables}: Dynamic identifiers for bound variables, where the token represents the variable's type rather than its name. For instance, in \texttt{Definition f a b c : nat := a + b + c}, the variables \texttt{a}, \texttt{b}, and \texttt{c} are tokenized based on their shared type \texttt{nat}, reflecting that their specific names are irrelevant to the semantic meaning.
    \item \textbf{Reserved Tokens}: Coq keywords, syntactic markers, hint databases, internal tactics, and special tokens we introduce for proof state representation (e.g., \texttt{\_Anonymous} for unnamed hypotheses, \texttt{goalcompleted} for solved goals, \texttt{REL} for De Bruijn indices).
\end{itemize}

\paragraph{Coverage and Limitations}
The tokenizer achieves 99.8\% coverage of Coq constructs in our corpus. The remaining 0.2\% falls back to default handling for two reasons: \begin{itemize} 
\item entities we intentionally exclude from extraction, such as primitive axioms and low-level type theory constructs that are not relevant to typical proofs
\item corner cases in Coq's syntax that our pattern matching does not capture. 
\end{itemize}
When encountering these cases, the tokenizer defaults to treating them as local variables with empty type information. This fallback strategy ensures robustness without impacting the quality of tokenization for mainstream Coq proofs.

This semantic tokenization enables models to learn from mathematical content rather than syntactic variations, providing the foundation for the Planner-Executor architecture described next.

\section{Examples of Semantic Information Types}
\label{sec:appendix_examples}
To clarify the different forms of semantic information used in our evaluation—\textbf{Origin}, \textbf{Internal}, and \textbf{Intuition}—we present two examples below. Each example corresponds to a formal concept from Coq's libraries, with its representation shown under the three semantic views.

\subsection{Example 1: Fixpoint Construction via \texttt{FixFun}}

This example uses the definition \texttt{TLC.LibFix.FixFun}, a fixed-point combinator to define recursive functions on types with inhabited codomain.

\paragraph{Origin:}
\begin{verbatim}
TLC.LibFix.FixFun A B {IB : Inhab B} (F : (A -> B) -> A -> B) : A -> B :=
  FixFunMod eq F
\end{verbatim}

\paragraph{Internal:}
\begin{verbatim}
fun ( A : Type ) =>
  fun ( B : Type ) =>
    fun ( IB : ( TLC.LibLogic.Inhab.Inhab B ) ) =>
      fun ( F : forall ( _Anonymous : forall ( _Anonymous : A ) -> B ) ->
                 forall ( _Anonymous : A ) -> B ) =>
        ( TLC.LibFix.FixFunMod A B IB ( Coq.Init.Logic.eq.eq B ) F )
\end{verbatim}

\paragraph{Intuition:}
\begin{quote}
This function provides a way to define recursive functions in a non-recursive language by finding a fixed point of a higher-order function, ensuring termination and correctness through the use of inhabited types and equality.
\end{quote}

\subsection{Example 2: Well-Founded Recursion via \texttt{FixWf}}

This example uses \texttt{Equations.Type.Subterm.FixWf}, a definition that supports well-founded recursion in dependent type theory.

\paragraph{Origin:}
\begin{verbatim}
Equations.Type.Subterm.FixWf
  `{WF : WellFounded A R}
  (P : A -> Type)
  (step : forall x : A, (forall y : A, R y x -> P y) -> P x)
  : forall x : A, P x :=
  Fix wellfounded P step
\end{verbatim}

\paragraph{Internal:}
\begin{verbatim}
fun ( A : Type ) =>
  fun ( R : Equations.Type.Relation.relation A ) =>
    fun ( WF : Equations.Type.Classes.WellFounded A R ) =>
      fun ( P : A -> Type ) =>
        fun ( step : forall ( x : A ) ->
                      forall ( y : A ) ->
                      forall ( _ : R y x ) -> P y -> P x ) =>
          Equations.Type.WellFounded.Fix
            A R
            (Equations.Type.Classes.wellfounded A R WF)
            P step
\end{verbatim}

\paragraph{Intuition:}
\begin{quote}
This theorem provides a way to define recursive functions on a type \texttt{A} with a well-founded relation \texttt{R}, ensuring that the recursion terminates by only allowing recursive calls on elements that are 'smaller' according to \texttt{R}. It is a foundational tool for defining well-behaved recursive functions in dependent type theory.
\end{quote}

\subsection{Summary}

These examples illustrate how semantic information about a Coq concept can be represented in different forms:

\begin{itemize}
    \item \textbf{Origin}: The original Coq source code, written by developers.
    \item \textbf{Internal}: A machine-level, compiler-oriented abstraction.
    \item \textbf{Intuition}: A natural language explanation that communicates conceptual meaning to humans.
\end{itemize}

These formats serve as distinct inputs to evaluate the level of understanding that a model comprehends formal concepts.

\section{Prompt Template Used for Coq Proof Generation Process}
\label{sec:appendix_prompt_format}

The following prompt template is employed to represent the current proof state and context in our experiments. This template captures essential elements such as hypotheses, goals, referenced global definitions, proof tracing history, related premises and tactics, curated notes, hints, and available user actions. It is designed to guide the model to either request additional information or suggest tactics for proof progression.

\begin{minipage}[t]{\textwidth}
\centering
\begin{lstlisting}[frame=single, 
                   basicstyle=\scriptsize\ttfamily,
                   columns=fixed,
                   keepspaces=true,
                   breaklines=true,
                   postbreak=\mbox{\textcolor{red}{$\hookrightarrow$}\space}]
I am currently working on a formal proof in Coq. Here is my current state and context:

=== Current Proof States ===
# Hypotheses:
{hyps}

# Goal:
{goal}

Global definitions referenced:
# Glob def:
{glob_def}
=== Proof Tracing ===
This shows how we reached the current state through previous tactics:

Tactics: {tactic_seq}
{proof_summary}

=== Related Premises ===
Potentially relevant premises (for reference only):
{premises}

=== Related Tactic ===
Commonly used tactics for similar proofstates (for reference only):
{tactics}

=== Public Notes ===
Curated insights relevant to current proof:
{public_notes}

=== Hint ===
Some hints may help you to understand the proof:
{hint}

=== Available Actions ===

Please choose ONE of the following actions:

1. Request more information about specific concepts/tactics mentioned above
Your response must be in this format:
{{
  "info": ["concept_name1", "concept_name2", "tactic1", "tactic2", ...]
}}

2. Suggest a list of up to 10 tactics to try - prefer single atomic tactics over compound ones unless the combination is highly confident. I will provide the compiler's response for each
Your response must be in this format:
{{
  tactics: [
    {{"tactic": "tactic1", "reason": "explanation for why this specific tactic is recommended"}},
    {{"tactic": "tactic2", "reason": "explanation for why this specific tactic is recommended"}},
    ...
  ]
}}
\end{lstlisting}
\end{minipage}

This prompt format provides a consistent interaction framework across all conditions. The different information configurations are instantiated by varying the content inserted into specific fields such as \texttt{glob\_def} and \texttt{public\_notes}.





\end{document}